\pdfoutput=1

\documentclass[11pt]{article}

\usepackage[final]{coling}

\usepackage{times}
\usepackage{latexsym}

\usepackage{algorithm}
\usepackage{algorithmic}
\usepackage{multirow}
\usepackage{graphicx}
\usepackage{listings}
\usepackage[normalem]{ulem}
\useunder{\uline}{\ul}{}

\lstdefinestyle{mystyle}{
    backgroundcolor=\color{white},
    commentstyle=\color{comment},
    keywordstyle=\color{keyword},
    numberstyle=\tiny\color{gray},
    stringstyle=\color{string},
    basicstyle=\ttfamily\footnotesize,
    breaklines=true,
    captionpos=b,
    numbers=left,
    numbersep=5pt,
}

\usepackage[T1]{fontenc}

\usepackage[utf8]{inputenc}

\usepackage{microtype}

\usepackage{inconsolata}

\title{\textit{Poetry in Pixels}: Prompt Tuning for Poem Image Generation via Diffusion Models}

\author{
Sofia Jamil{$^1$} \enspace Bollampalli Areen Reddy{$^1$} \enspace Raghvendra Kumar{$^1$} \\ \enspace \textbf{ Sriparna Saha{$^1$}} \enspace \textbf{ K J Joseph{$^2$}} \enspace 
\textbf{ Koustava Goswami{$^2$}} \\
{{$^1$} {Department of Computer Science \& Engineering, Indian Institute of Technology Patna, India}}  \quad\\
{{$^2$} {Adobe Research} } \\
$^1$\{sofia\_2321cs16, bollampalli\_2101cs19, raghvendra\_2221cs27, sriparna\}@iitp.ac.in \\
$^2$\{koustavag, josephkj\}@adobe.com
}

\begin{document}
\maketitle
\begin{abstract}

The task of text-to-image generation has encountered significant challenges when applied to literary works, especially poetry. Poems are a distinct form of literature, with meanings that frequently transcend beyond the literal words. 
To address this shortcoming, we propose a \textbf{\textit{PoemToPixel}} framework designed to generate images that visually represent the inherent meanings of poems. Our approach incorporates the concept of prompt tuning in our image generation framework to ensure that the resulting images closely align with the poetic content. In addition, we propose the \textbf{\textit{PoeKey}} algorithm, which extracts three key elements in the form of emotions, visual elements, and themes from poems to form instructions which are subsequently provided to a diffusion model for generating corresponding images. Furthermore, to expand the diversity of the poetry dataset across different genres and ages, we introduce \textbf{\textit{MiniPo}}, a novel multimodal dataset comprising 1001 children's poems and images. Leveraging this dataset alongside \textit{PoemSum}, we conducted both quantitative and qualitative evaluations of image generation using our \textit{PoemToPixel} framework. This paper demonstrates the effectiveness of our approach and offers a fresh perspective on generating images from literary sources. \textit{The code and dataset used in this
work are publicly available.\footnote{\url{https://github.com/SofeeyaJ/Poetry-In-Pixels-Coling2025}}}.

\end{abstract}

\section{Introduction}

Poetry, with its layered meanings and expressive language, often paints descriptive images in the mind of the reader. But what if these imagined scenes could be brought to real vision? Visualizing poetry adds a new dimension to the experience, allowing us to see and feel the poem in a way that extends beyond the words. In recent years, the automatic generation of realistic images from arbitrary text descriptions has attracted a lot of attention. Several initiatives have been made to address the zero-shot text-to-image generation challenge by pre-training large-scale generative models on massive datasets of image-text pairs, such as DALL-E \cite{dalle} and CogView \cite{cogview}. Unlike typical text, creating images for poems presents a significant challenge, as the deeper meanings embedded in poetry can easily be overlooked if only the literal interpretation is considered. However, with the advancements in these models, an important question arises: \emph{Can expressive forms of language, such as poetry, be visually represented in a way that accurately captures their intended meaning? To address this, we explore the potential of Diffusion Models and Large Language Models (LLMs) in the domain of poetry. We propose the \textbf{\textit{PoemToPixel}} pipeline, which is built on the summarization capabilities of LLMs along with the image synthesis capabilities of Diffusion Models.}


Poems frequently challenge the conventional narrative structures found in stories and paragraphs, prompting a unique approach to analysis and interpretation \cite{introo1}. Taking into consideration this attribute of poetry, our \textit{PoemToPixel} framework adopts a two-step strategy for poem visualization, combining poem summarization and poem key element extraction. For summarization, we use the GPT-4o mini\cite{gpt4} Model, which has shown impressive results across a variety of summarization tasks \cite{2gpt,1gpt}. \emph{Despite these favourable outcomes, our analysis revealed limitations in the summaries generated by GPT for creative content, highlighting the need for further refinement. As a result, we employed Prompt Tuning based on human feedback on the GPT-4o mini model to improve the quality of poem summaries.} These generated summaries enhance comprehension by transforming abstract concepts into more concrete insights, making the poet's message clearer and more engaging. With these textual summaries available, they can be transformed into images that vividly convey the ideas of the poem. However, the complexity of crafting effective prompts poses a significant challenge for the average user. This difficulty arises from the major difference between user-provided natural language prompts and the keyword-enriched prompts required for the system's high-quality output \cite{intro1, intro2, intor3}. \emph{In order to bridge this gap, we propose the \textbf{\textit{PoeKey}} algorithm, which extracts \textbf{KEY} elements from \textbf{POE}m summaries. These key elements extracted from the poems are then formulated into precise instructions, which are fed into diffusion models for image generation. Furthermore, refining user language into system language is crucial for enhancing the user experience \cite{prompt2, openai2023, intro1}. With this in mind, we repeatedly modified prompts by utilizing a human feedback loop that incorporated both quantitative and qualitative measures. Based on this feedback, the prompts were further changed to achieve better image generation.} To summarize, we make the following key contributions: 

\begin{itemize}
    \item We present a novel approach of poem-to-image generation through our innovative \textbf{\textit{PoemToPixel}} pipeline, a framework designed to produce multimodal outputs in the form of visual images.
    \item We introduce \textit{\textbf{PoeKey}}, a novel algorithm designed to extract three key elements from poems, facilitating the creation of high-quality prompts for image generation.
    \item We implement a two-step prompt tuning process to analyze the summarization capability of GPT and the image generation capability of diffusion models. 
    \item We introduce a novel multimodal dataset \textit{\textbf{MiniPo}} consisting of 1001 children's rhymes that integrate textual and visual elements, along with metadata, thereby enhancing the available resources for poem analysis across diverse age groups.
\end{itemize}

\section{Related Works}

\textbf{Text-to-Image Prompting:}
With the emergence of Large Language Models (LLMs), the need for efficient prompting has become more apparent. 
However, in tasks like text-to-image generation, the pre-training of LLMs does not fully address how prompts influence the image creation process. Automatically refining user input into system-optimized prompts is a crucial step toward developing more user-friendly text-to-image systems \cite{xie2023prompt}. Some approaches include interactive systems that assist users by suggesting immediate improvements to help refine their prompts, though manual adjustments are still often necessary \cite{feng2023promptmagician, brade2023promptify, liu2022design}. \emph{To address this, our \textit{PoemToPixel} approach combines human-guided prompt tuning with the interactive capabilities of LLMs to generate prompts that effectively drive text-to-image models, producing relevant images for poetry.}
\newline\textbf{Image and Poetry:} There has been limited research focused on bridging the gap between poetry and images. A few studies have focused on generating poems from images using a deep coupled visual-poetic embedding model combined with RNN-based adversarial training \cite{poemfromimage1}. In the study by \cite{poemfromimage2}, poetry was generated from images using recurrent neural networks trained on existing poems. By contrast, \cite{li2021paint4poem} worked on generating images from poems, specifically focusing on creating paintings that replicated Feng Zikai's unique artistic style, representing a distinct form of art altogether. However, current literature lacks approaches for generating images from poems, irrespective of genre, type, or style. \emph{Our work addresses this gap by introducing the PoemToPixel approach, which captures the complex meanings within poems, resulting in relevant image generation.}
\newline \textbf{ Diffusion Models:} 
The field of text-to-image generation has seen significant advancements, particularly with the introduction of various diffusion model techniques \cite{diff2,diff3,diff4,diff10,diff11}. These models have proven to be highly effective at generating realistic and diverse images from textual prompts. However, as highlighted by \cite{ramesh2022hierarchical}, diffusion models often struggle with complex prompts that require advanced skills, such as binding attributes to objects and spatial reasoning. Recent works \cite{diff12,diff13} have aimed to enhance the capabilities of pre-trained diffusion models to better capture intricate details in textual descriptions. Despite these improvements, the models still fall short in accurately representing the finer details in longer, more complex prompts. \emph{Our PoemToPixel approach addresses this challenge by formulating prompts that convey intricate details in a concise and precise manner, improving the generation of relevant images from poetry.}




\section{Corpus}
In the existing literature, the datasets that are available for poetry generation are predominantly unimodal (text-based). To accomplish our research objectives, we employed two datasets.
\begin{table}[!ht]
\centering
\resizebox{\columnwidth}{!}{%
\begin{tabular}{lcc}
\hline
\textit{Type} & \textit{MiniPo} & \textit{PoemSum} \\ \hline
\textit{Number of Poems} & 1001 & 3011 \\
\textit{Max Poem Length (in words)} & 346 & 6830 \\
\textit{Avg. Poem Length (in words)} & 50.47 & 209 \\
\textit{Max Summary Length (in words)} & 491 & 1104 \\
\textit{Avg. Summary length (in words) } & 105.79 & 141 \\ \hline
\end{tabular}%
}
\caption{Statistics of the two Datasets that we have used in our research.}
\label{dataset table}
\end{table}

\textit{\textbf{PoemSum:}}
The first dataset, \textit{PoemSum} \cite{data1}, contains 3011 poetry samples and their English-language summaries. In the \textit{PoemSum} dataset, the poems are collected from multiple online sources, whereas the corresponding standard reference summaries were obtained from `Poem Analysis' \footnote{\url{https://poemanalysis.com/}}, a website known for its comprehensive archive of high-quality poem summaries. 

\textit{\textbf{MiniPo:}} To facilitate and promote poetry research, we introduce a novel dataset,  \textit{`MiniPo'}, consisting of 1001 samples that address the lack of representation of children's poetry in the existing dataset. Existing datasets covered various genres but did not include any children's poems. \textit{`MiniPo'} aims to enhance diversity in poem analysis, allowing for a more comprehensive and inclusive study of poetry across different age groups and genres. Each sample in the dataset includes the poem's title, the poem itself, and images depicting the poem's plot. We will now outline the steps taken to create our dataset.
\newline \textbf{1) Data Collection:}
The {\em MiniPo} dataset was curated from various online open sources, with a specific focus on nursery rhymes.
\footnote{\url{https://www.nurseryrhymes.org/nursery-rhymes.html
}} \footnote{\url{https://smart-central.com/} }. 
The choice to focus on nursery poems was inspired by their concise structure and simple meanings, making them ideal for both present and future poetry research. 

To ensure the reliability and accuracy of the data collection process, a group of three English-proficient final-year undergraduate students were selected. The selection criteria ensured that these individuals possessed the necessary technical skills to verify the collected data, specifically poems sourced from various online websites.
\newline \textbf{2) Data Cleaning: }
After collecting data from various sources, additional processing was required to ensure dataset consistency. We extracted textual content from nursery rhymes and presented it as metadata to our dataset for future manipulation. The refined records were then verified to ensure that any inconsistencies were addressed.
We have provided different
statistics of the \textit{PoemSum} and \textit{MiniPo} dataset in Table \ref{dataset table}.







\section{Proposed Methodology}

\subsection{Problem Statement}


Given a dataset \( \mathcal{D} = \{P_1, P_2, \ldots, P_n\} \), where each data point \( P_i \) represents a poem, the objective is to generate a corresponding set of images \( \{I_{P1}, I_{P2}, \ldots, I_{Pn}\} \) such that each image \( I_{P_i} \) visually represents the interpretation of the poem \( P_i \). The textual poems \( P_i \) are summarized to \( S_i \) using a summarization module, \( S_i = f_{summ}(P_i) \), where \( f_{summ} \) represents the summarization function of the LLM. Leveraging the summary \( S_i \), key components of poems such as themes, emotions, and visual elements are extracted, \( E_i = f_{KeyExtraction}(S_i) \), utilizing the poem embeddings and similarity scores. The extracted key components \( E_i \) are then fed into an image generation framework \( \mathcal{F} \) to produce the images \( I_{P_i} \), \( I_{P_i} = \mathcal{F}(E_i) \), where \( \mathcal{F} \) is the image generation function applied to the key components. The aim is to ensure that each generated image \( I_{P_i} \) accurately reflects the visual tone of its corresponding poem \( P_i \).

\subsection{\textit{PoemToPixel} Framework}

To make poetry more understandable and visually appealing to readers, we introduce the \textit{PoemToPixel} framework, which converts textual poems into visual illustrations. In this section, we present our \textit{PoemToPixel} pipeline, which allows for training-free image generation that accurately captures the meanings embedded in the verses. The overall procedure of our method is divided into the following phases:





\subsubsection{Phase 1: Summarization Module}

\begin{figure}[!ht]
\centerline{\includegraphics[width=0.9\columnwidth]{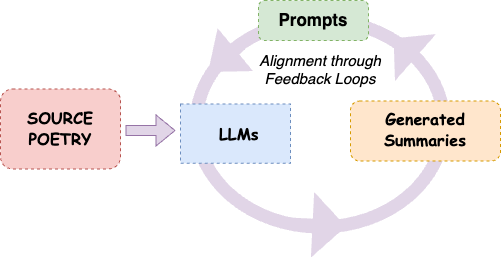}}
\caption{A framework of an iterative process with prompts refined based on feedback for improved summarization.}
\label{summmodule}
\end{figure}

The complex nature of poetic language presents a unique set of challenges during summarization. Poems often contain rich, multifaceted language, such as metaphors, similes, and symbolic imagery, making it difficult to capture their essence in a concise summary without losing the intended meaning or emotional impact. The emergence of Large Language Models (LLMs) such as GPT creates novel possibilities for innovation in this field. While GPT has demonstrated strong performance in a variety of text summarization tasks, like long document summarization \cite{longdocumentsummariation}, dialogue summarization \cite{dialogue_summarisation}, and news summarization \cite{news_summaration}, its capabilities in the genre of poetry are largely unexplored. To address this, we prompted GPT-4o mini model to summarize poems. However, the quality of GPT’s output is highly dependent on the input prompts \cite{antar}, making the design of effective prompts important. As shown in Figure \ref{summmodule}, we implemented prompt tuning based on feedback, aligning our prompts through both qualitative (human feedback) and quantitative measures. The refinement process for these prompts is detailed below, and we repeated the process iteratively based on the results of these evaluations.
\newline \textbf{Prompt Tuning through Human Feedback:} Automatic evaluation metrics can often be insufficient for summarization tasks, particularly in the context of poetry, where meaning is deeply embedded and comprehension is varied. To address this issue, we collaborated with a team of four professional experts from the Poetry Society of the Nation, which included one renowned poet, two university professors, and one PhD in visual and literary arts. We chose 10\% of the samples from the \textit{PoemSum} dataset for prompt tuning. We asked the experts to rate the summaries generated by GPT-4o mini, assigning either a +1 reward or a -1 penalty based on the quality of the summaries compared to gold reference summaries. The process was repeated until the overall score decreased after any round. Table \ref{exper_summ_scores} shows the total scores after each round of summary generation, given the particular prompt. Based on expert feedback, we refined the prompts for positive scores by modifying keywords, focusing on specific details, or highlighting key aspects. To maintain fairness and avoid bias against AI-generated summaries, we did not inform the experts that the summaries were generated by GPT. After six rounds of iterative feedback, the score decreased in the seventh round, after which we discontinued the process. We concluded that the sixth prompt \( \mathcal{R}_6\) was the most effective for poem summarization using GPT-4o mini.

\( \mathcal{R}_6\): \textit{Summarize below, covering the main theme, mood, and any notable literary devices used by the poet.}
\newline \textit{All the subsequent prompts used in our Prompt tuning through feedback are listed in the Appendix \ref{tuning_summarisation}, with the total expert scores provided in Table \ref{exper_summ_scores}}.
\newline \textbf{Prompt Tuning through Automated Evaluation: } Apart from human evaluation, automated evaluation is often used for its efficiency and reliability. Most summarization tasks are based on metrics like ROUGE \cite{rouge1}, BLEU \cite{bleu1}, and METEOR \cite{meteor} scores. For our automatic evaluation, we used ROUGE and METEOR scores to assess the performance of summaries generated by various prompts, as shown in Table \ref{exper_summ_scores}. The results indicate that Round 6 achieved impressive scores in both metrics. However, the differences in scores among the seven prompts within the same model were relatively minor. Based on these findings, we conducted further experiments on the entire dataset using the finalized prompt \( \mathcal{R}_6\).

\begin{table}[!ht]
\centering
\resizebox{\columnwidth}{!}{%
\begin{tabular}{lccc}
\hline
 & \multicolumn{1}{l}{\textit{Expert Scores}} & \multicolumn{1}{l}{\textit{Rouge - L}} & \multicolumn{1}{l}{\textit{Meteor}} \\ \hline
\textit{Prompt 1} & 0 & 0.1720 & 0.1488 \\
\textit{Prompt 2} & 2 & 0.1717 & 0.1564 \\
\textit{Prompt 3} & 2 & 0.1770 & 0.1827 \\
\textit{Prompt 4} & 4 & 0.1811 & 0.1825 \\
\textit{Prompt 5} & 4 & \textbf{0.1821} & 0.1855 \\
\textit{Prompt 6} & 4 & 0.1801 & \textbf{0.2140} \\
\textit{Prompt 7} & 2 & 0.1779 & 0.2041 \\ \hline
\end{tabular}%
}
\caption{Sum of Expert Scores provided for summarization based on quality of summaries generated in each round in comparison to gold reference summaries.}
\label{exper_summ_scores}
\end{table}

\subsubsection{Key Element Extraction Unit}


We implement our proposed \textit{PoeKey} algorithm to extract emotions, visual elements, and themes from poems, which are necessary for shaping and guiding the visual representation of the poem. Emotions within the poems are identified using an emotion classifier \cite{emotion_citation} that assigns an emotion tag based on the most dominant sentiment depicted in the poem summary. Additionally, visual elements are identified by utilizing nouns, pronouns, and named entities described within the poem summaries. For theme extraction, we compute the sentence embedding of the poem summary using Sentence-BERT \cite{sentence_bert}. Cosine similarity is then calculated between the poem's embedding and each predefined theme embedding, with the theme corresponding to the highest similarity score assigned to the poem. These themes, which commonly recur across various literary works and genres, were identified using relevant online resources \footnote{\url{https://www.poetryfoundation.org/topics-themes}}
\footnote{\url{https://poemanalysis.com/poetry-explained/poetry-themes/}} that highlight major themes in poetry. Additionally, we collaborated with literary scholars specializing in literature to create concise descriptions for each theme. 


\subsubsection{Instruction Generator}

\begin{figure}[!ht]
\centerline{\includegraphics[width=0.8\columnwidth]{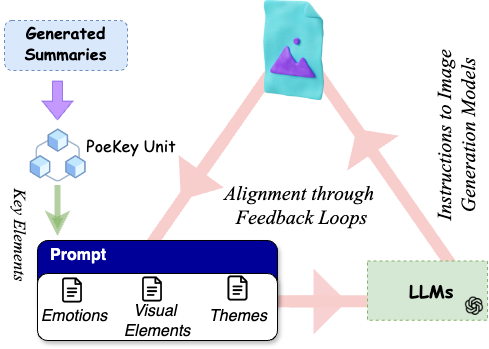}}
\caption{A framework of an iterative process of image prompts refined based on feedback.}
\label{imagegen}
\end{figure}

The key components extracted from the poems are then provided to an instruction generation module. This module incorporates GPT-4o mini and the SDXL Turbo diffusion model \cite{sdxl_turbo}. We selected the SDXL Turbo diffusion model for its cost-effective and time-efficient image generation process. As shown in Figure \ref{imagegen}, the process begins by providing GPT-4o mini with a prompt and the extracted elements (Visual Elements, Themes, and Emotions) as the context. The output is an instruction for creating an image corresponding to a particular poem, which will be fed to a diffusion model for image generation.


\textbf{Prompt to GPT}: \emph{Create a prompt which can generate an image which will be able to represent the poem using the given emotion and visual elements and theme of the poem, and the prompt should be under 50 words.}


\textbf{Instruction by GPT}: \emph{To convey the emotion of sadness in the context of mortality and the afterlife, I envision an image that captures a dense jungle setting, symbolizing the overwhelming presence of life and death intertwined. The focal point would be a solitary parrot perched on a gnarled branch.}

\begin{figure}[!ht]
\centerline{\includegraphics[width=0.9\columnwidth]{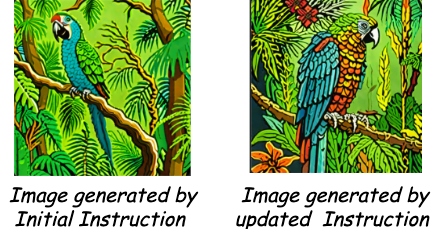}}
\caption{An instance of image generation phases during instruction tuning.}
\label{expert}
\end{figure}

The diffusion model then generates an image based on the instructions. We evaluated the images both automatically and manually described below, and we improved our prompts based on feedback.
\newline \textbf{Human Evaluation}
Assessing the quality of generated images for poems requires human evaluations, as the images must capture the embedded meanings and emotions of the poetry. To conduct this evaluation, we collaborated with the aforementioned four area experts.
We selected 5\% of the samples from the \textit{PoemSum} dataset. The experts reviewed the images produced by the diffusion model to identify whether they captured the emotion, themes, and meaning of the poems. Each expert rated the images on a scale of 1 to 5, and the average score was calculated after each round. A higher score indicated better alignment with the poems, whereas a lower score indicated misalignment. Based on expert feedback, we refined the GPT prompts, which in turn adjusted the instruction prompts and generated images based on those instructions as demonstrated in Figure \ref{expert}.

\textbf{Updated Prompt to GPT:} \emph{Write a prompt for an image that visually represents the poem's mood and themes using the given emotional tone and visual details. Keep the prompt concise, under 50 words.}

\textbf{Updated Instruction by GPT: } \textit{Create an image of a vibrant parrot perched on a gnarled branch in a dense, shadowy jungle. Surround it with wilted plants and hints of decay, embodying a sense of waiting and longing, reflecting mortality and the somber afterlife.}

This iterative process resulted in a gradual improvement in the coherence of the images and poems as outlined in Table \ref{exper_image_scores}. After five rounds of feedback, the average score began to decline in the sixth round, at which point we stopped the process and finalized the fifth prompt. 

\( \mathcal{I}_5\): \textit{Formulate a prompt to generate an image that reflects the poem's emotional depth and themes using the specified visual elements, keeping the prompt within 50 words.}
\newline \textit{All the prompts used for prompt tuning in image generation are shown in the Appendix \ref{tuning_image_generation}, along with the average expert scores outlined in Table \ref{exper_image_scores}}.
\newline \textbf{Automated Evaluation: }
Along with human evaluation, we also performed automated evaluation to reduce the time and cost involved in manual expert review. For this, we used two well-established quantitative metrics: Image-Text Matching Loss (ITM) \cite{itmitc} and Image-Text Contrastive Loss (ITC) \cite{itmitc}. The results shown in Table \ref{exper_image_scores} demonstrate that the \( \mathcal{I}_5\) prompt achieved the highest scores in both metrics. Based on these results, we proceed with the image generation task on the entire dataset using this finalized prompt.

\begin{table}[!ht]
\centering
\resizebox{\columnwidth}{!}{%
\begin{tabular}{lccc}
\hline
 & \multicolumn{1}{l}{\textit{Expert Scores}} & \multicolumn{1}{l}{\textit{ITM}} & \multicolumn{1}{l}{\textit{ITC}} \\ \hline
\textit{Instruction 1} & 1.75 & 0.111 & 0.221 \\
\textit{Instruction 2} & 3 & 0.330 & 0.246 \\
\textit{Instruction 3} & 3 & 0.333 & 0.272 \\
\textit{Instruction 4} & 3.75 & 0.310 & 0.269 \\
\textit{Instruction 5} & 4.25 & \textbf{0.356} & \textbf{0.276}\\
\textit{Instruction 6} & 4 & 0.334 & 0.263 \\ \hline
\end{tabular}%
}
\caption{Expert scores provided for image generation based on alignment of images with respect to poems.}
\label{exper_image_scores}
\end{table}




\section{Experiment and Results}


\subsection{Phase 1: Summarization}
\textbf{Comparison with the Baselines:} The experiments were conducted using GPT-4o mini and included comparisons with popular summarization models such as T5 \cite{T5}, BART \cite{lewis2019bart}, Pegasus \cite{pegasus}, and Falcon \cite{falcon} on both datasets. In recent years, these models have demonstrated impressive performance across various summarization tasks \cite{laskar2022domain, ravaut2022summareranker}. For the \textit{PoemSum} dataset, we evaluated the model-generated summaries against the gold reference summaries using a range of established metrics, including ROUGE-1 (R1), ROUGE-2 (R2), ROUGE-L (RL), BLEU-1, BLEU-2, BLEU-3, BLEU-4, and METEOR, as shown in Table \ref{quantsummaries}. These metrics collectively provide a comprehensive assessment of the accuracy, fluency, and overall quality of the generated summaries across all models. 


\begin{table}[!ht]
\centering
\resizebox{\columnwidth}{!}{%
\begin{tabular}{lccccrc}
\hline
\multicolumn{1}{l|}{Scores} & \multicolumn{4}{c|}{Baselines} & \multicolumn{2}{c}{GPT-4o mini} \\ \hline
 & \textit{Pegasus} & \textit{Bart} & \textit{T5} & \textit{Falcon} & \multicolumn{1}{c}{\textit{Without PT}} & \textit{With PT} \\ \hline
\textit{R1} & 0.0167 & 0.2006 & 0.1915 & 0.2065 & 0.2921 & \textbf{0.3188} \\
\textit{R2} & 0.0015 & 0.0421 & 0.0356 & 0.0374 & 0.0428 & \textbf{0.0586} \\
\textit{RL} & 0.0141 & 0.1343 & 0.1279 & 0.1321 & 0.1720 & \textbf{0.1801} \\
\textit{BLEU-1} & 0.0495 & 0.1837 & 0.1812 & 0.2313 & 0.5327 & \textbf{0.6541} \\
\textit{BLEU-2} & 0.0153 & 0.1323 & 0.1324 & 0.1706 & 0.3986 & \textbf{0.5107} \\
\textit{BLEU-3} & 0.0067 & 0.0729 & 0.0703 & 0.0916 & 0.2318 & \textbf{0.3102} \\
\textit{BLEU-4} & 0.0036 & 0.0419 & 0.0389 & 0.0502 & 0.1376 & \textbf{0.1867} \\
\textit{METEOR} & 0.0111 & 0.1041 & 0.0998 & 0.1089 & 0.1488 & \textbf{0.2140} \\ \cline{1-7}
\end{tabular}%
}
\caption{Quantitative evaluation of generated summaries from different models on the \textit{PoemSum} Dataset using Different approaches. Here, PT refers to Prompt Tuning with Feedback Loops}
\label{quantsummaries}
\end{table}

\textbf{Key Observations:} The results presented in Table \ref{quantsummaries} demonstrate that summaries produced using GPT-4o mini with Prompt Tuning outperformed those generated by all base models, including GPT-4o mini models without Prompt Tuning. Furthermore, as shown in Table \ref{summarisation}, applying Prompt Tuning improved the generated summaries, making them more reflective of the poem's underlying meaning and more similar to the gold reference summaries. In contrast, summaries generated without Prompt Tuning failed to capture the true interpretation of the poem. Additionally, the base models were unable to provide meaningful summaries, often producing identical verses with minimal variation.
\begin{table*}[!ht]
\centering
\resizebox{\textwidth}{!}{%
\begin{tabular}{llll}
\hline
\multicolumn{1}{c}{\textit{\textbf{Poem}}} & \multicolumn{1}{c}{\textit{\textbf{Reference Summary}}} & \multicolumn{1}{c}{\textit{\textbf{Baseline (Falcon) Summary}}} & \multicolumn{1}{c}{\textit{\textbf{GPT}}} \\ \hline
\textit{\begin{tabular}[c]{@{}l@{}}Lovely dainty Spanish needle\\ With your yellow flower and white,\\ Dew bedecked and softly sleeping,\\ Do you think of me to-night?\\ Shadowed by the spreading mango,\\ Nodding o'er the rippling stream,\\ Tell me, dear plant of my childhood,\\ Do you of the exile dream?\\ Do you see me by the brook's side\\ Catching crayfish 'neath the stone,\\ As you did the day you whispered:\\ Leave the harmless dears alone?\\ Do you see me in the meadow\\ Coming from the woodland spring\\ With a bamboo on my shoulder\\ And a pail slung from a string?\\ Do you see me all expectant\\ Lying in an orange grove,\\ While the swee-swees sing above me,\\ Waiting for my elf-eyed love?\\ Lovely dainty Spanish needle,\\ Source to me of sweet delight,\\ In your far-off sunny southland\\ Do you dream of me to-night?\end{tabular}} & \begin{tabular}[c]{@{}l@{}}The Spanish Needle' by Claude McKay \\ is a beautiful, nostalgic \\ poem that looks back on a speaker's \\ childhood. In the stanzas, \\ the speaker asks the Spanish needle, \\ a type of plant, if it remembers him during \\ his youth. He has fond \\ memories of spending time outside,\\ near a river, carrying water, and spending \\ time with his love. It's clear he's feeling \\ nostalgic about his past and hoping that \\ through the plant, a tiny bit of himself \\ still exists in the past.\end{tabular} & \begin{tabular}[c]{@{}l@{}}Lovely dainty Spanish needle\\  With your yellow flower and white, \\ Dew bedecked and softly sleeping, \\ Do you think of me to-night? \\ Shadowed by the spreading mango, \\ Nodding o'er the rippling stream, \\ Tell me, dear plant of my childhood . \\ Do you see me in the meadow \\ Coming from the woodland spring \\ With a bamboo on my shoulder \\ And a pail slung from a string?\end{tabular} & \begin{tabular}[c]{@{}l@{}} \textbf{ \textit{1) Prompt Tuning through Feedback: }} \\ In this nostalgic poem, the speaker addresses a \\ dainty Spanish needle flower, reminiscing \\ about their childhood adventures and \\ longing for a connection with the past. \\ The emotional tone is one of wistfulness \\ and yearning, as the speaker questions \\ whether the flower remembers them \\ and dreams of their joyful moments together. \\ Ultimately, the poem reflects a deep sense\\ of nostalgia and a desire for companionship, \\ even across distance and time.\\ \\ \textbf{ \textit{2) Without Prompt Tuning:}} \\ The poem reflects nostalgia and longing \\ through the perspective of a person reminiscing \\ about their childhood in a sunny, southern land. \\ The speaker addresses a Spanish needle plant, \\ reminiscing about their past experiences and \\ childhood activities such as catching crayfish, \\ fetching water, and waiting for a loved one. \\ The plant, with its yellow and white flowers, \\ is seen as a symbol of the speaker’s lost \\ homeland and past joys.\end{tabular} \\ \hline
\end{tabular}%
}
\caption{Sample of the Generated Summary by the Baseline (Falcon) model and GPT-4o mini Model after Prompt Tuning through Feedbacks with respect to gold reference summaries.}
\label{summarisation}
\end{table*}

\subsection{Phase 2: Image Generation}

\begin{figure*}[!htbp]
\centerline{\includegraphics[width=\textwidth]{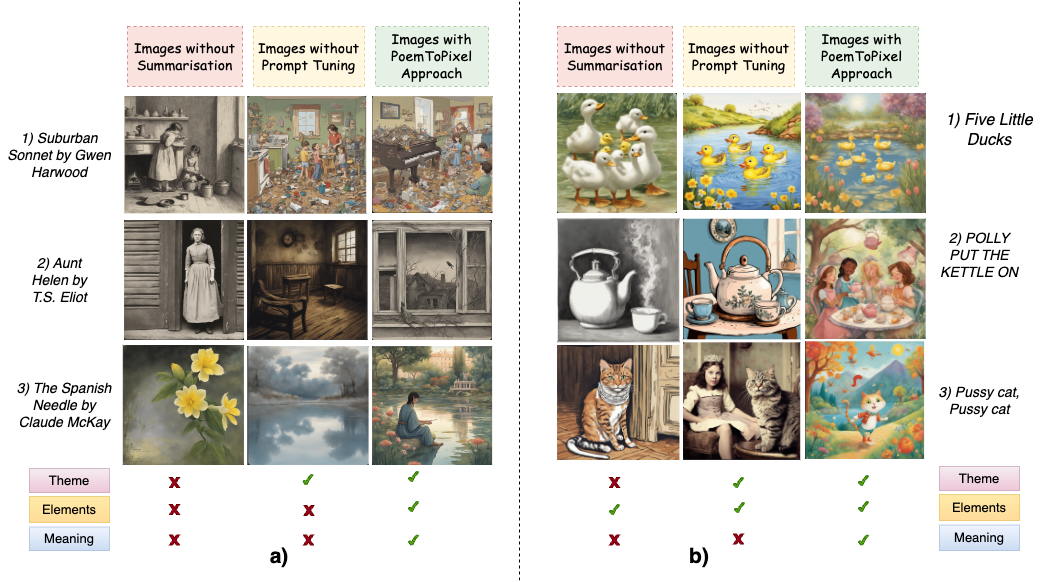}}
\caption{A comparison of Poem to Image generation using SDXL Base Diffusion Model with different methods and \textit{PoemToPixel} Approach on (a) \textit{PoemSum} dataset (b) \textit{MiniPo} ((a collection of Nursery Rhymes) dataset.}
\label{poemtale-example}
\end{figure*}

\textbf{Qualitative Evaluation: }
Figure \ref{poemtale-example} presents a detailed comparison of images produced by different approaches for various poems using SDXL Base 1.0 \cite{SDXL}. \textit{The complete poems mentioned here are provided in the Appendix \ref{_appendixcompletepoems}}. Our qualitative analysis focused on three key aspects: \newline \textbf{1) Theme: }The \textit{PoemToPixel} Pipeline demonstrated a remarkable ability to capture the emotional theme and setting of poetry verses. As shown in Figure \ref{poemtale-example} (a), without the summarization approach, the images generated struggled to convey the intended theme or sentiment of the poem. In contrast, when using the summarization approach, the emotional tone of the poem was accurately reflected in the generated images. Similarly, for our proposed \textit{MiniPo} dataset, as shown in Figure \ref{poemtale-example} (b), images generated without summarization did not reflect the poem's setting as depicted in the verses, whereas images generated with summarization successfully captured the joyous theme (for instance, Poem 2 and Poem 3 in Figure \ref{poemtale-example} (b)).
\newline \textbf{2) Visual Elements: } The \textit{PoemToPixel} Framework demonstrated an exceptional ability to capture the visual aspects of the poems due to the incorporation of our proposed \textit{PoeKey} algorithm, which extracted the visual elements described in the poem. For instance, in Figure \ref{poemtale-example} (b), Poem 3 illustrates its capacity to depict aspects such as flowers, rivers, and the act of sitting by the riverside, precisely representing the imagery described in the poem. In contrast, the images produced without  \textit{PoemToPixel} framework failed to capture these essential visual elements, as shown in Poem 3 in Figure \ref{poemtale-example}(a). \textbf{3) Meaning: }Figures \ref{poemtale-example} (a) and \ref{poemtale-example} (b) demonstrate the \textit{PoemToPixel} Framework's potential to efficiently comprehend and convey the deeper meaning of the poems. This effectiveness is primarily due to the summarization process, which captures the complex meanings present in the verses, leading to the formation of precise instructions for image generation models. In contrast, images generated without prompt tuning or summarization failed to visually represent the poem's meaning. For example, in Figure \ref{poemtale-example} (b), for the famous nursery rhyme ``Polly, Put the Kettle On," the \textit{PoemToPixel} framework accurately captures the essence of ``everyone drinking tea," as well as the key element—the kettle. On the other hand, other approaches only managed to depict the kettle, missing the broader context of the poem.
\newline \textbf{Quantitative Evaluation: }
The results presented in Table \ref{image_quant} provide valuable insights into the performance of various image generation approaches, highlighting the advantages of each method in the context of our task. The key findings include:
\newline \textbf{1) \textit{PoemToPixel}: } Our proposed approach outperformed all other methods, achieving the highest ITM (Image-Text Matching) and ITC (Image-Text Coherence) scores for both datasets. This superior performance can be attributed to our two-phase image generation process, in which summarization effectively captured the poem's meaning, and our proposed \textit{PoeKey} algorithm extracted the core elements of the poem to incorporate into the generated images. This combination greatly improved the alignment of the poem and its visual representation.
\textbf{2) Performance of Different Approaches: } Besides the \textit{PoemToPixel} method, we compared two other approaches: (i) generating images directly from the poem without summarization, and (ii) generating images without prompt tuning, where the image models received an initial instruction without any feedback mechanism. The results indicate that image generation models, when directly provided with the poem text, faced challenges grasping the intricate and hidden meanings within the verses, leading to irrelevant images. Moreover, without prompt tuning, the models underperformed compared to those with feedback, showing a significant improvement of 34.97\% in ITM and 27.4\% in ITC scores when using prompt tuning. This highlights the importance of feedback mechanisms in producing accurate and contextually relevant images.
\begin{table}[!ht]
\centering
\resizebox{\columnwidth}{!}{%
\begin{tabular}{lcccc}
\hline
Dataset & Scores & Without summary & Without PT & With \textit{PoemToPixel} \\ \hline
\multicolumn{1}{c|}{\multirow{2}{*}{PoemSum}} & ITM & 0.123 & 0.251 & \textbf{0.386} \\
\multicolumn{1}{c|}{} & ITC & 0.205 & 0.274 & \textbf{0.312} \\ \cline{1-1}
\multicolumn{1}{l|}{\multirow{2}{*}{MiniPo}} & ITM & 0.333 & 0.508 & \textbf{0.618} \\
\multicolumn{1}{l|}{} & ITC & 0.306 & 0.329 & \textbf{0.364} \\ \hline
\end{tabular}%
}
\caption{Quantitative Comparisons of Image Generation Approach for Poems on \textit{PoemSum} and \textit{MiniPo} Dataset using SDXL Base Diffusion Model. Here, PT referes to Prompt Tuning.}
\label{image_quant}
\end{table}
\newline \textbf{3) Dataset-Based Performance: } We conducted image generation experiments on two datasets, \textit{PoemSum} and \textit{MiniPo}, which varied in size, genre, form, and style. As shown in Table \ref{image_quant}, the \textit{PoemToPixel} pipeline achieved 37.54\% higher ITM and 14.28\% higher ITC scores in the \textit{MiniPo} dataset than in the \textit{PoemSum} dataset. This is possibly due to the simpler themes of nursery rhymes in the \textit{MiniPo} dataset, in contrast to the more complex meanings found in the poems present in the \textit{PoemSum} dataset. Additionally, the shorter length of \textit{MiniPo} poems (an average of 58 words) compared to \textit{PoemSum} (an average of 209 words) made it easier for the model to generate relevant imagery. Despite these advantages in the \textit{MiniPo} dataset, other methods still struggled to produce images that aligned with the rhymes, whereas our \textit{PoemToPixel} approach delivered strong results across both datasets.

\section{Conclusion}
In conclusion, the \textit{PoemToPixel} framework makes a valuable contribution to the field of poetic image generation. It demonstrates the ability to interpret the intricate meanings embedded in poems and generate corresponding visuals. Our \textit{PoeKey} algorithm further enhances this by accurately capturing the emotions, themes, and visual elements necessary for an authentic representation of poetic imagery. Additionally, we introduce the \textit{MiniPo} Dataset, consisting of 1001 nursery rhymes, each paired with images generated by the \textit{PoemToPixel} framework, establishing a new multimodal dataset for children's rhymes. Looking ahead, we aim to expand our research by generating multiple images that depict different segments of a single poem.

\section{Limitations}
While the \textit{PoemToPixel} framework effectively captures the meanings, emotions, and visual elements in poems, it faces challenges when multiple emotions or meanings are expressed within a single poem, as one image cannot encompass all these aspects. Additionally, the framework lacks language-agnostic capabilities, as it currently only supports English and does not extend to other languages.

\section{Ethical Consideration}

A key ethical consideration involves the inherent biases present in diffusion models, which may reflect societal, cultural, or data-driven biases from the pre-trained models. These biases can potentially influence the generation of images related to poems on specific topics or forms, resulting in unfair or inappropriate outputs.
\bibliography{custom}

\appendix

\section{Appendix}
\label{sec:appendix}

\subsection{\textit{MiniPo} Dataset}

\begin{figure}[!htbp]
\centerline{\includegraphics[width=0.9\columnwidth]{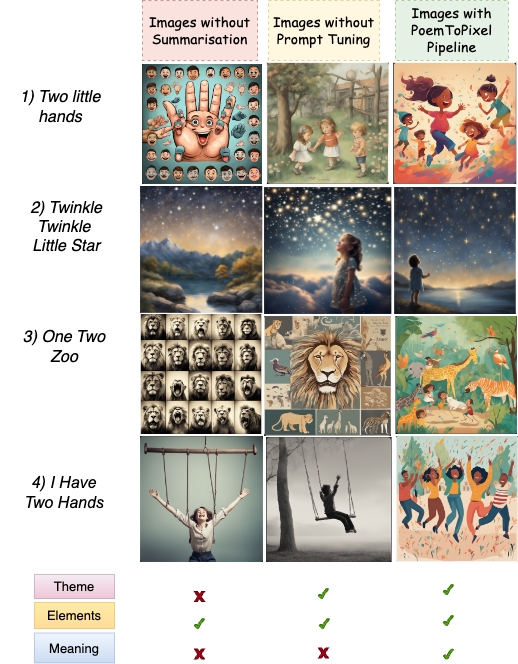}}
\caption{A comparison of poem to image generation using SDXL Base Diffusion Model for poem mentioned in Table \ref{minipo_poems_only} with different methods and \textit{PoemToPixel} Approach}
\label{appendix_poem_tale}
\end{figure}

\begin{figure}[!htbp]
\centerline{\includegraphics[width=0.9\columnwidth]{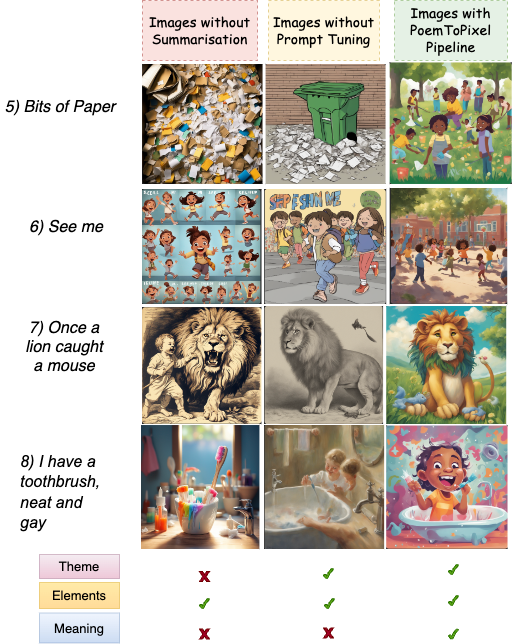}}
\caption{A comparison of poem to image generation using SDXL Base Diffusion Model for poem mentioned in Table \ref{minipo_poems_only} with different methods and \textit{PoemToPixel} Approach}
\label{appendix_poem_tale2}
\end{figure}

This section presents instances of poems in Table \ref{minipo_poems_only} from our proposed multimodal \textit{MiniPo} dataset, along with the corresponding images in Figure \ref{appendix_poem_tale} and Figure \ref{appendix_poem_tale2} generated by our \textit{PoemToPixel} framework.

\begin{table}[!ht]
\centering
\resizebox{\columnwidth}{!}{

\begin{tabular}{ll}
\hline
\multicolumn{2}{c}{\textbf{MiniPo Dataset}} \\ \hline
\textbf{Title} & \multicolumn{1}{c}{\textit{\textbf{Poems}}} \\ \hline
{\ul 1) Two little hands} & \textit{\begin{tabular}[c]{@{}l@{}}Two little hands go Clap clap clap\\ Two little feet goTap tap tap\\ Two little eyes That open wide\\ One little head Nods side to side\end{tabular}} \\ \hline
\begin{tabular}[c]{@{}l@{}}2) Twinkle Twinkle \\ Little Star\end{tabular} & \textit{\begin{tabular}[c]{@{}l@{}}Twinkle, twinkle little star, how I wonder what you are. \\ Up above the world so high, like a diamond in the sky. \\ Twinkle, twinkle little star, how I wonder what you are.\end{tabular}} \\ \hline
{\ul 3) One Two Zoo} & \textit{\begin{tabular}[c]{@{}l@{}}One, two, one, two lets visit the zoo\\ Three, four, three, four hear the lion roar\\ Five, six, five, six  watch the monkey tricks\\ Seven, eight, seven, eight peacock looks so great\\ Nine, ten, nine, ten we shall come again.\end{tabular}} \\ \hline
{\ul 4) I Have Two hands} & \textit{\begin{tabular}[c]{@{}l@{}}I have two arms that swing like this, (thrice) \\ I have two arms that swing like this, Swing, Swing, Swing.\\ I have two hands that clap like this,(thrice) \\ I have two hands that clap like this, Clap, Clap, Clap.\\ I have little fingers that wriggle like this, (thrice)\\ I have little fingers that wriggle like this, Wriggle, Wriggle, Wriggle.\\ I have two legs that jump like this, (thrice) \\ I have two legs that jump like this, Jump, Jump, Jump.\\ I have two feet that march like this, (thrice) \\ I have two feet that march like this, March, March, March.\\ I have little toes that hop like this, (thrice) \\ I have little toes that hop like this,Hop, Hop, Hop.\end{tabular}} \\ \hline
{\ul 5) Bits of Paper} & \textit{\begin{tabular}[c]{@{}l@{}}Bits of paper ,Bits of paper\\ Lying on the floor, Lying on the floor\\ Makes the place untidy, Makes the place untidy\\ Pick them up, Pick them up\\ Collect all the papers, Collect all the papers\\ Where shall we throw?Where shall we throw?\\ Throw them in the dustbin, Throw them in the dustbin\\ Now the place is clean, Now the place is clean\end{tabular}} \\ \hline
{\ul 6) See me} & \textit{\begin{tabular}[c]{@{}l@{}}See me skip, see me run,\\ I am going to school Like everyone.\\ See me shout, See me wave,\\ I am off to school Bye, everyone.\\ See me hug, See me dance,\\ I see my friends Hello, everyone!\\ See me smile, See me grin,\\ When the bell rings I walk in. \\ See me jump, See me play,\\ I am in S.P.M. Public SchoolHip Hip Hooray!!!\end{tabular}} \\ \hline
{\ul 7) Once a Lion caught a mouse} & \textit{\begin{tabular}[c]{@{}l@{}}Once a lion caught a mouse. The mouse begged the lion not to\\ eat him. Someday, he said, he would help the lion. The lion set\\ the mouse free.\\ \\ One day, the lion got caught in a hunter’s net. The mouse came\\ to help lion. He cut the net with his sharp teeth. The lion was set\\ free. The lion and the mouse became good friends.\end{tabular}} \\ \hline
{\ul 8) I have a tooth brush neat and gay} & \textit{\begin{tabular}[c]{@{}l@{}}I have a toothbrush, neat and gay,\\ To brush my teeth with, everyday.\\ I brush them each morning,\\ \\ Then I brush them each night;\\ Till all are shining, Clean and bright.\end{tabular}} \\ \hline
\end{tabular}
}
\caption{List of Poems of MiniPo Datset}
\label{minipo_poems_only}
\end{table}

\subsection{\textit{PoemSum} Dataset}

\begin{figure}[!htbp]
\centerline{\includegraphics[width=0.9\columnwidth]{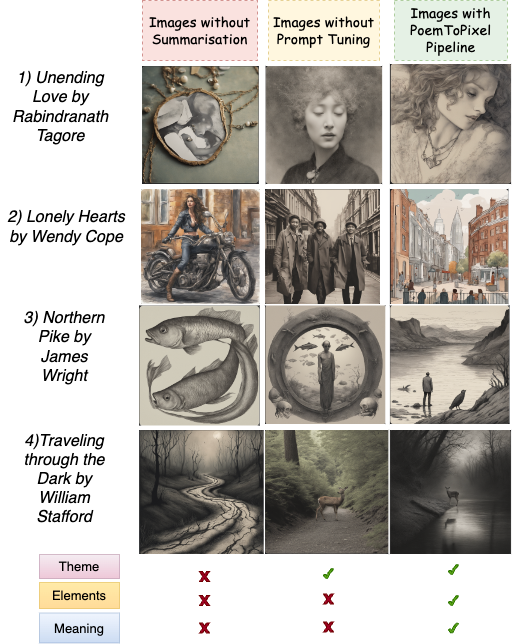}}
\caption{A comparison of poem to image generation using SDXL Base Diffusion Model for poem mentioned in Table \ref{poemsum_poems_only} with different methods and \textit{PoemToPixel} Approach}
\label{appendix_poem_sum_images}
\end{figure}

This section presents instances of poems in Table \ref{poemsum_poems_only} from \textit{PoemSum} dataset, along with the corresponding images in Figure \ref{appendix_poem_sum_images} generated by our \textit{PoemToPixel} framework.

\begin{table}[!ht]
\centering
\resizebox{\columnwidth}{!}{%
\begin{tabular}{ll}
\hline
\multicolumn{2}{c}{\textbf{PoemSum Dataset}} \\ \hline
\textbf{Title} & \multicolumn{1}{c}{\textit{\textbf{Poems}}} \\ \hline
{\ul 1) Unending Love by Rabindranath Tagore} & \textit{\begin{tabular}[c]{@{}l@{}}I seem to have loved you in numberless forms, numberless times…\\ In life after life, in age after age, forever.\\ My spellbound heart has made and remade the necklace of songs,\\ That you take as a gift, wear round your neck in your many forms,\\ In life after life, in age after age, forever.\\ \\ Whenever I hear old chronicles of love, its age-old pain,\\ Its ancient tale of being apart or together.\\ As I stare on and on into the past, in the end you emerge,\\ Clad in the light of a pole-star piercing the darkness of time:\\ You become an image of what is remembered forever.\\ \\ You and I have floated here on the stream that brings from the fount.\\ At the heart of time, love of one for another.\\ We have played alongside millions of lovers, shared in the same\\ Shy sweetness of meeting, the same distressful tears of farewell-\\ Old love but in shapes that renew and renew forever.\\ \\ Today it is heaped at your feet, it has found its end in you\\ The love of all man’s days both past and forever:\\ Universal joy, universal sorrow, universal life.\\ The memories of all loves merging with this one love of ours –\\ And the songs of every poet past and forever.\end{tabular}} \\ \hline
{\ul 2) Lonely Hearts by Wendy Cope} & \textit{\begin{tabular}[c]{@{}l@{}}Can someone make my simple wish come true?\\ Male biker seeks female for touring fun.\\ Do you live in North London? Is it you?\\ Gay vegetarian whose friends are few,\\ I'm into music, Shakespeare and the sun.\\ Can someone make my simple wish come true?\\ Executive in search of something new—\\ Perhaps bisexual woman, arty, young.\\ Do you live in North London? Is it you?\\ Successful, straight and solvent? I am too—\\ Attractive Jewish lady with a son.\\ Can someone make my simple wish come true?\\ I'm Libran, inexperienced and blue—\\ Need slim, non-smoker, under twenty-one.\\ Do you live in North London? Is it you?\\ Please write (with photo) to Box 152.\\ Who knows where it may lead once we've begun?\\ Can someone make my simple wish come true?\\ Do you live in North London? Is it you?\end{tabular}} \\ \hline
{\ul 3) Northern Pike by James Wright} & \textit{\begin{tabular}[c]{@{}l@{}}All right. Try this,\\ Then. Every body\\ I know and care for,\\ And every body\\ Else is going\\ To die in a loneliness\\ I can't imagine and a pain\\ I don't know. We had\\ To go on living. We\\ Untangled the net, we slit\\ The body of this fish\\ Open from the hinge of the tail\\ To a place beneath the chin\\ I wish I could sing of.\\ I would just as soon we let\\ The living go on living.\\ An old poet whom we believe in\\ Said the same thing, and so\\ We paused among the dark cattails and prayed\\ For the muskrats,\\ For the ripples below their tails,\\ For the little movements that we knew \\ the crawdads were making\\ under water,\\ \\ For the right-hand wrist of my cousin who is a policeman.\\ We prayed for the game warden's blindness.\\ We prayed for the road home.\\ We ate the fish.\\ There must be something very beautiful in my body,\\ I am so happy.\end{tabular}} \\ \hline
{\ul 4) Traveling through the Dark by William Stafford} & \textit{\begin{tabular}[c]{@{}l@{}}Traveling through the dark I found a deer\\ dead on the edge of the Wilson River road.\\ It is usually best to roll them into the canyon:\\ that road is narrow; to swerve might make more dead.\\ By glow of the tail-light I stumbled back of the car and stood by the heap, \\ a doe, a recent killing; she had stiffened already, almost cold.\\ I dragged her off; she was large in the belly.\\ My fingers touching her side brought me the reason—\\ her side was warm; her fawn lay there waiting, alive, \\ still, never to be born.\\ Beside that mountain road I hesitated.\\ The car aimed ahead its lowered parking lights; \\ under the hood purred the steady engine.\\ I stood in the glare of the warm exhaust turning red; \\ around our group I could hear the wilderness listen.\\ I thought hard for us all—my only swerving—, \\ then pushed her over the edge into the river.\end{tabular}} \\ \hline
\end{tabular}%
}
\caption{List of Poems of PoemSum Datset}
\label{poemsum_poems_only}
\end{table}

\subsection{Prompt Tuning for Summarisation}
\label{tuning_summarisation}

Summaries were generated using different prompts and compared to the gold reference summaries from the \textit{PoemSum} dataset. Experts evaluated the quality of the generated summaries against the gold references, assigning a +1 for high quality and -1 for low quality. The expert scores for each round are presented in the Table \ref{appendix_scores_summ} below.

\begin{table*}[!ht]
\centering
\resizebox{\textwidth}{!}{%
\begin{tabular}{llccccc}
\hline
 & \textit{\textbf{Prompts}} & \multicolumn{1}{l}{\textit{\textbf{Score 1}}} & \multicolumn{1}{l}{\textit{\textbf{Score 2}}} & \multicolumn{1}{l}{\textit{\textbf{Score 3}}} & \multicolumn{1}{l}{\textit{\textbf{Score 4}}} & \multicolumn{1}{l}{\textit{\textbf{\begin{tabular}[c]{@{}l@{}}Total \\ Score\end{tabular}}}} \\ \hline
\textit{\( \mathcal{R}_1\)} & \textit{Summarize the following poem.} & 1 & -1 & 1 & -1 & 0 \\ \hline
\textit{\( \mathcal{R}_2\)} & \textit{\begin{tabular}[c]{@{}l@{}}Summarize the following poem in 2-3 sentences, \\ focusing on the central theme and mood.\end{tabular}} & 1 & 1 & -1 & 1 & 2 \\ \hline
\textit{\( \mathcal{R}_3\)} & \textit{\begin{tabular}[c]{@{}l@{}}Summarize the poem in 2-3 sentences, \\ ensuring to convey the poet’s emotional tone \\ and the underlying message.\end{tabular}} & 1 & 1 & -1 & 1 & 2 \\ \hline
\textit{\( \mathcal{R}_4\)} & \textit{\begin{tabular}[c]{@{}l@{}}Summarize the following poem, \\ focusing on the central theme and mood.\end{tabular}} & 1 & 1 & 1 & 1 & 4 \\ \hline
\textit{\( \mathcal{R}_5\)} & \textit{\begin{tabular}[c]{@{}l@{}}Summarize below, covering the main emotion, \\ and  mood of the poem.\end{tabular}} & 1 & 1 & 1 & 1 & 4 \\ \hline
\textit{\( \mathcal{R}_6\)} & \textit{\begin{tabular}[c]{@{}l@{}}Summarize below, covering the main theme, \\ mood, and any notable literary devices used by the poet.\end{tabular}} & 1 & 1 & 1 & 1 & 4 \\ \hline
\textit{\( \mathcal{R}_7\)} & \textit{\begin{tabular}[c]{@{}l@{}}Summarize the poem below, \\ capturing the poet’s emotional tone, \\ theme, mood and the underlying message.\end{tabular}} & 1 & 1 & 1 & -1 & 2 \\ \hline
\end{tabular}%
}
\caption{Expert scores at the end of each round to mark the summaries generated by GPT, comparing them to the gold standard summaries using the provided prompts \( \mathcal{R}_i\).}
\label{appendix_scores_summ}
\end{table*}

\begin{table*}[!ht]
\centering
\resizebox{\textwidth}{!}{%
\begin{tabular}{llccccc}
\hline
 & \textit{\textbf{Prompts}} & \multicolumn{1}{l}{\textit{\textbf{Score 1}}} & \multicolumn{1}{l}{\textit{\textbf{Score 2}}} & \multicolumn{1}{l}{\textit{\textbf{Score 3}}} & \multicolumn{1}{l}{\textit{\textbf{Score 4}}} & \multicolumn{1}{l}{\textit{\textbf{\begin{tabular}[c]{@{}l@{}}Total \\ Score\end{tabular}}}} \\ \hline
\textit{\( \mathcal{I}_1\)} & \textit{\begin{tabular}[c]{@{}l@{}}Create a prompt which can generate an image \\ which will be able reperesent the poem using the \\ given emotion and visual elemnts and theme of the poem \\ and the prompt should be under 50 words \end{tabular}} & 2 & 2 & 1 & 2 & 1.75 \\ \hline
\textit{\( \mathcal{I}_2\)} & \textit{\begin{tabular}[c]{@{}l@{}}Create a prompt to generate an image t\\ hat captures the poem’s essence by combining \\ the specified emotion, visual elements, and theme, \\ ensuring the prompt is concise and under 50 words.\end{tabular}} & 3 & 3 & 2 & 4 & 3 \\ \hline
\textit{\( \mathcal{I}_3\)} & \textit{\begin{tabular}[c]{@{}l@{}}Write a prompt for an image that visually represents\\  the poem's mood and themes using the given emotional \\ tone and visual details. Keep the prompt concise, under 50 words.\end{tabular}} & 3 & 3 & 3 & 3 & 3 \\ \hline
\textit{\( \mathcal{I}_4\)} & \textit{\begin{tabular}[c]{@{}l@{}}Design a prompt that generates an image \\ capturing the essence of the poem through a combination \\ of the specified emotions, visual elements, and thematic undertones.\end{tabular}} & 3 & 4 & 4 & 4 & 3.75 \\ \hline
\textit{\( \mathcal{I}_5\)} & \textit{\begin{tabular}[c]{@{}l@{}}Formulate a prompt to generate an image \\ that reflects the poem's emotional depth and themes \\ using the specified visual elements, keeping the prompt \\ within 50 words.\end{tabular}} & 3 & 5 & 4 & 5 & 4.25 \\ \hline
\textit{\( \mathcal{I}_6\)} & \textit{\begin{tabular}[c]{@{}l@{}}Compose a prompt capable of generating an image \\ that visually interprets the poem, emphasizing the \\ given emotions, key visual elements, and thematic elements.\end{tabular}} & 3 & 4 & 4 & 5 & 4 \\ \hline
\end{tabular}%
}
\caption{Expert scores at the end of each round for the images generated based on the instructions provided by GPT through these prompts \( \mathcal{I}_i\).}
\label{appendix_scores_image}
\end{table*}

\subsection{Prompt Tuning for Image Generation}
\label{tuning_image_generation}

Images were generated for 5\% of the \textit{PoemSum} dataset using various instructions produced by GPT in response to input prompts, as shown in the Table \ref{appendix_scores_image} . Experts evaluated the images based on how effectively they conveyed the meaning of the poems, assigning scores on a scale of 1 to 5. Based on the total scores, the input prompts were refined, leading to updated instructions from GPT, which were then used to regenerate the images.
\label{prompt_image_appendix}







\subsection{Description of evaluation metrics used for summarization}

ROUGE-1 measures the overlap of unigrams, while ROUGE-2 evaluates bigram overlap, and ROUGE-L assesses the longest common subsequence between the generated and reference texts, focusing on both sequence similarity and fluency. The BLEU scores assess the precision of n-gram overlaps, with BLEU-1 for unigrams, BLEU-2 for bigrams, BLEU-3 for trigrams, and BLEU-4 for four-grams. Additionally, METEOR computes the alignment of unigrams and synonym matches between the generated and reference texts, incorporating precision, recall, and stemming to assess overall translation quality.

\subsection{Poems used in Qualitative Analysis of Poems}
\label{_appendixcompletepoems}

The complete poems, referenced by their titles in Figures \ref{poemtale-example} (a) and (b), are listed in the Table \ref{completepoems} below.

\begin{table*}[!ht]
\centering
\resizebox{\textwidth}{!}{%
\begin{tabular}{lclc}
\hline
\multicolumn{2}{c|}{\textbf{\begin{tabular}[c]{@{}c@{}}PoemSum Dataset \\ (Figure a)\end{tabular}}} & \multicolumn{2}{c}{\textbf{\begin{tabular}[c]{@{}c@{}}MiniPo Dataset\\ (Figure b)\end{tabular}}} \\ \hline
\multicolumn{1}{c}{\textit{\textbf{Title}}} & \textit{\textbf{Poems}} & \textbf{Title} & \textit{\textbf{Poems}} \\ \hline
\begin{tabular}[c]{@{}l@{}}1) Suburban Sonnet \\ by Gwen Harwood\end{tabular} & \textit{\begin{tabular}[c]{@{}c@{}}She practises a fugue, though it can matter\\ to no one now if she plays well or not.\\ Beside her on the floor two children chatter,\\ then scream and fight. She hushes them. A pot\\ boils over. As she rushes to the stove\\ too late, a wave of nausea overpowers\\ subject and counter-subject. Zest and love\\ drain out with soapy water as she scours\\ the crusted milk. Her veins ache. Once she played\\ for Rubinstein, who yawned. The children caper\\ round a sprung mousetrap where a mouse lies dead.\\ When the soft corpse won't move they seem afraid.\\ She comforts them; and wraps it in a paper\\ featuring: Tasty dishes from stale bread.\end{tabular}} & {\ul 1) Five Little Ducks} & \textit{\begin{tabular}[c]{@{}c@{}}Five little ducks went swimming one day\\ Over the hill and far away\\ Mother duck said, “Quack quack quack quack”\\ And only four little ducks came back!\\ \\ Four little ducks went swimming one day\\ Over the hill and far away\\ Mother duck said, “Quack quack quack quack”\\ And only three little ducks came back!\\ \\ Three little ducks went swimming one day\\ Over the hill and far away\\ Mother duck said, “Quack quack quack quack”\\ And only two little ducks came back!\\ \\ Two little ducks went swimming one day\\ Over the hill and far away.\\ Mother duck said, “Quack quack quack quack”\\ And only one little duck came back!\\ \\ One little duck went swimming one day\\ Over the hill and far away\\ Mother duck said, “Quack quack quack quack”\\ And all her five little ducks came back!\end{tabular}} \\ \hline
\begin{tabular}[c]{@{}l@{}}2) Aunt Helen \\ by T.S. Eliot\end{tabular} & \textit{\begin{tabular}[c]{@{}c@{}}Miss Helen Slingsby was my maiden aunt,\\ And lived in a small house near a fashionable square\\ Cared for by servants to the number of four.\\ Now when she died there was silence in heaven\\ And silence at her end of the street.\\ The shutters were drawn and the undertaker wiped his feet—\\ He was aware that this sort of thing had occurred before.\\ The dogs were handsomely provided for,\\ But shortly afterwards the parrot died too.\\ The Dresden clock continued ticking on the mantelpiece,\\ And the footman sat upon the dining-table\\ Holding the second housemaid on his knees—\\ Who had always been so careful while her mistress lived.\end{tabular}} & {\ul 2) POLLY PUT THE KETTLE ON} & \textit{\begin{tabular}[c]{@{}c@{}}Polly, put the kettle on,\\ Polly, put the kettle on, Polly, put the kettle on, We'll all have tea.\\ Sukey, take it off again,\\ Sukey, take it off again,\\ Sukey, take it off again,\\ They've all gone away.\\ Polly, put the kettle on,\\ Polly, put the kettle on,\\ Polly, put the kettle on, We'll all have tea.\\ Blow the fire and make the toast, \\ Put the muffins on to roast, \\ Blow the fire and make the toast, We'll all have tea.\\ Polly, put the kettle on,\\ Polly, put the kettle on,\\ Polly, put the kettle on, We'll all have tea.\\ Sukey, take it off again,\\ Sukey, take it off again,\\ Sukey, take it off again,\\ They've all gone away.\end{tabular}} \\ \hline
\begin{tabular}[c]{@{}l@{}}3) The Spanish Needle \\ by Claude McKay\end{tabular} & \textit{\begin{tabular}[c]{@{}c@{}}Lovely dainty Spanish needle\\ With your yellow flower and white,\\ Dew bedecked and softly sleeping,\\ Do you think of me to-night?\\ Shadowed by the spreading mango,\\ Nodding o'er the rippling stream,\\ Tell me, dear plant of my childhood,\\ Do you of the exile dream?\\ Do you see me by the brook's side\\ Catching crayfish 'neath the stone,\\ As you did the day you whispered:\\ Leave the harmless dears alone?\\ Do you see me in the meadow\\ Coming from the woodland spring\\ With a bamboo on my shoulder\\ And a pail slung from a string?\\ Do you see me all expectant\\ Lying in an orange grove,\\ While the swee-swees sing above me,\\ Waiting for my elf-eyed love?\\ Lovely dainty Spanish needle,\\ Source to me of sweet delight,\\ In your far-off sunny southland\\ Do you dream of me to-night?\end{tabular}} & {\ul 3)Pussy Cat Pussy Cat} & \textit{\begin{tabular}[c]{@{}c@{}}Pussy cat, pussy cat,\\ \\ Where have you been?\\ \\ "I've been to London to\\ \\ Look at the Queen."\\ \\ \\ Pussy cat, pussy cat,\\ \\ What did you there?\\ \\ "I frightened a little mouse\\ \\ Under the chair."\end{tabular}}
\end{tabular}%
}
\caption{Complete list of Poems mentioned in Figure \ref{poemtale-example}.}
\label{completepoems}
\end{table*}

\end{document}